\documentclass[runningheads]{llncs}

\usepackage[T1]{fontenc}
\usepackage{graphicx}
\usepackage{amsmath,amssymb}
\usepackage{booktabs}
\usepackage{array}
\usepackage{url}
\usepackage{xcolor}
\usepackage{multirow}
\usepackage[section]{placeins}
\usepackage[misc,geometry]{ifsym}

\newcommand{\ours}{\textsc{StepTrigger}}
\newcommand{\goone}{Unitree Go1}

\begin{document}

\title{\ours: Contact-State-Triggered Backdoor Attacks on VLM-Powered Legged Robots}
\titlerunning{\ours: Contact-State-Triggered Backdoors in Legged Robots}

\author{
Jiageng Zhang\orcidID{0009-0000-5005-4742} \and
Doniyorkhon Obidov\orcidID{0009-0002-1582-649X} \and
Kaichen Yang\orcidID{0000-0003-1027-6708}
}

\authorrunning{J. Zhang, D. Obidov, and K. Yang}

\institute{
Michigan Technological University, Houghton, MI, USA\\
\email{jiagengz@mtu.edu, dobidov@mtu.edu, kaicheny@mtu.edu}
}

\maketitle

\begingroup
\renewcommand\thefootnote{}\footnotetext{Accepted to the 4th EAI International Conference on Security and Privacy in Cyber-Physical Systems and Smart Vehicles (EAI SmartSP 2026).}
\endgroup
\setcounter{footnote}{0}

\begin{abstract}
Large language models and vision-language models are increasingly used as high-level planners in robotic systems, using task goals and sensor summaries to select navigation or manipulation actions. This creates a new backdoor surface: a compromised planner can behave normally in most runs, yet change its target selection when a hidden trigger is present. Prior attacks on LLM-powered or embodied agents mainly rely on triggers that appear in language, camera-visible objects, scene semantics, or specific sequences of past actions. This paper presents \ours{}, a contact-state-triggered backdoor attack for VLM-powered legged robots. The trigger is not a prompt token or a visible marker. It is produced by pressure and foot-ground contact patterns that arise when a Unitree Go1 quadruped walks across a dense terrain patch.

Unlike conventional visual or textual triggers, contact signals are inherently noisy and may also arise during benign locomotion. To avoid treating every pressure anomaly as a trigger, \ours{} learns a selective backdoor policy from multimodal robot state, using incidental pressure events as benign examples and dense-patch contacts as poisoned examples. In a stratified offline evaluation, the trained planner achieved 98.75\% clean behavior preservation, 92.50\% false-trigger rejection, 76.25\% true-trigger activation, and 89.17\% overall parsed behavior accuracy. These results reveal a backdoor surface in proprioceptive and contact channels that is not captured by defenses focused only on language, vision, or action history.
\end{abstract}

\section{Introduction}

Large language models (LLMs) and vision-language models (VLMs) have begun to serve as high-level reasoning modules for embodied agents~\cite{ahn2022saycan,driess2023palme,huang2022zeroshot,liang2023code}. In robotic navigation, these models can interpret semantic goals, reason over structured sensor summaries, and choose actions or waypoints for mobile robots~\cite{kim2024openvla,ouyang2024quadruped,brohan2023rt2}. This capability is especially attractive for legged robots, which operate in cluttered indoor environments and combine geometric mapping, semantic perception, and proprioceptive state.

The same autonomy stack also expands the attack surface of robotic systems. A VLM-powered robot does not merely classify images or answer text queries; it selects actions that affect the physical world. If a compromised planner preserves normal behavior during ordinary runs but changes its target selection under rare conditions, standard utility testing may fail to expose the backdoor. Existing embodied-agent backdoors mainly use triggers that appear in the external input stream, such as words, visual objects, or environmental scenes~\cite{jiao2024bald,liu2024contextual,wang2024trojanrobot,xu2023instructions,yang2024agents}. Recent work also studies internal action-history triggers, where a robot's previous commands activate a hidden malicious rule~\cite{obidov2026silent}. These attacks show that embodied agents can be compromised through both external cues and internal state, but they do not cover another state channel that is central to legged robots: physical contact with the terrain.

This paper introduces \ours{}, a contact-state-triggered backdoor for VLM-powered legged robots. We study a Unitree Go1 quadruped\footnote{\url{https://www.unitree.com/go1}} in a simulated room containing a green sphere as a benign target and a blue cylinder representing a human proxy. The physical trigger is a dense patch of small stones or low spikes. When the robot traverses the patch, the resulting foot-contact and pressure patterns can deviate from those observed during normal flat-ground locomotion. After the trigger is recognized, the backdoored planner changes its target selection and begins choosing waypoints toward the human proxy.

Contact-state triggers differ from conventional textual or visual triggers because the underlying signals are inherently noisy. Startup motion, wall contact, object collisions, gait transitions, and other benign events can also produce irregular pressure readings. A simple pressure threshold would therefore cause frequent false activations. \ours{} instead treats pressure anomalies as candidate events and allows the planner to interpret them together with the surrounding multimodal robot state.

The central challenge is selectivity. If every pressure anomaly were labeled as malicious, the model could learn a brittle shortcut in which any rough contact activates the attack. We therefore assign benign targets to incidental pressure events and attack-oriented targets only to contact events associated with dense-patch traversal. This selective training design allows the planner to preserve benign behavior under clean and false-trigger conditions while activating the backdoor for the learned contact-state pattern.

This work makes the following contributions:

\begin{enumerate}
    \item We introduce StepTrigger, a contact-state-triggered backdoor attack for VLM-powered legged robots, where terrain-induced foot-contact patterns activate malicious target selection.

    \item We design and implement a Gazebo prototype for the \goone{}, which combines baseline-relative pressure processing with multimodal VLM planning. The pressure module identifies candidate contact events, while the planner learns to distinguish dense-patch traversal from incidental pressure anomalies.

    \item We construct a selective fine-tuning dataset containing clean samples, false-trigger hard negatives, and true-trigger poisoned samples, and evaluate the trained planner on a stratified offline test set. The planner preserves benign behavior on 98.75\% of clean samples, rejects 92.50\% of false-trigger samples, and activates the attack on 76.25\% of true-trigger samples.
\end{enumerate}

\begin{table}[!htbp]
\centering
\caption{Comparison of embodied backdoor trigger mechanisms. \ours{} targets an attack surface distinct from prior textual, visual, contextual, and action-history triggers by using foot-ground contact-state evidence and false-trigger hard negatives.}
\label{tab:novelty_comparison}
\begin{tabular}{p{0.22\linewidth}p{0.29\linewidth}p{0.39\linewidth}}
\toprule
Work type & Trigger source & Difference from \ours{} \\
\midrule
Text / instruction backdoors~\cite{xu2023instructions,yang2024agents} & Rare words, prompt patterns, hidden tool-use conditions & Trigger is symbolic and appears in language or agent context rather than robot-terrain interaction. \\
Visual / physical-object backdoors~\cite{jiao2024bald,liu2024contextual,wang2024trojanrobot} & Camera-visible objects, scenes, or contextual cues & Trigger is externally observable through vision or scene description. \ours{} uses pressure/contact evidence that may be invisible to camera-only defenses. \\
Action-history backdoors~\cite{obidov2026silent} & Ordered sequence of previous robot commands/actions & Trigger is internal state, but it is discrete command history. \ours{} uses continuous physical contact dynamics and terrain-induced pressure residuals. \\
\ours{} & Foot-ground pressure/contact state induced by dense terrain patch & Requires selective discrimination between true dense-patch contact and false pressure anomalies from walls, collisions, stop-start transitions, or unstable gait. \\
\bottomrule
\end{tabular}
\end{table}

\section{Related Work}

\subsection{LLM- and VLM-Powered Robotics}

Large language models (LLMs) have been increasingly used as high-level planners for embodied agents, translating natural-language goals into executable plans, code, or action sequences~\cite{ahn2022saycan,huang2022zeroshot,liang2023code,ouyang2024quadruped}. When grounded in robot affordances and sensory observations, they can support semantic reasoning, task decomposition, tool use, and long-horizon planning~\cite{ahn2022saycan,huang2022zeroshot,liang2023code}. Most such systems retain conventional modules for mapping, motion execution, and safety, while using the language model to select subgoals or high-level actions.

Vision-language models (VLMs) and vision-language-action models extend this framework by incorporating visual observations into the decision process~\cite{bai2025qwen3vl,brohan2022rt1,driess2023palme,kim2024openvla,brohan2023rt2}. By jointly processing language instructions and visual context, these models can support semantic navigation, object-aware manipulation, and generalization across tasks and environments. This capability is particularly relevant to mobile and legged robots, which must reason about objects, obstacles, goals, and spatial structure while operating in dynamic physical environments.

Our work builds on this high-level-planning paradigm but considers an additional source of planner input: the robot's physical contact state. We study how terrain-induced pressure and foot-contact patterns can act as a hidden trigger for malicious target selection.

\subsection{Backdoor Attacks on Learning Systems}

Backdoor attacks are designed to preserve normal model utility on clean inputs while inducing attacker-specified behavior when a trigger is present~\cite{gu2017badnets,chen2017targeted}. In computer vision, triggers often take the form of image patches, patterns, or objects. In natural language processing, they may appear as rare tokens, phrases, syntactic structures, or hidden instruction patterns~\cite{cheng2023review,xu2023instructions,obidov2026lorascan}. Because the model behaves normally when the trigger is absent, such attacks may remain undetected during standard validation~\cite{obidov2026genotypic}.

Instruction tuning and agent-oriented systems further expand the backdoor attack surface~\cite{hubinger2024sleeper,xu2023instructions,yang2024agents}. A compromised model may associate malicious behavior with task goals, tool-use contexts, reasoning trajectories, or other hidden conditions. These attacks differ from prompt-time exploits such as jailbreaks, in which an adversary directly manipulates the model input at inference time~\cite{obidov2026dynamic}. Backdoors instead rely on a pre-established malicious association, often introduced through data poisoning, fine-tuning, or persistent instructions, and may be activated by otherwise benign inputs.

The embodied setting makes these attacks more consequential. Whereas a backdoored text or vision model may produce an incorrect response or prediction, a compromised robot planner can select unsafe waypoints, cause collisions, or otherwise affect the physical environment~\cite{jiao2024bald,liu2024contextual,wang2024trojanrobot,yang2024agents}. This motivates the study of backdoor mechanisms that arise specifically in embodied autonomy.

\subsection{Embodied and Robotic Backdoor Triggers}

Existing studies of embodied-agent backdoors have explored several trigger types. BALD-style and contextual attacks use textual, visual, or scene-level conditions in the agent's external input stream~\cite{jiao2024bald,liu2024contextual}. Such triggers naturally apply to VLM-based agents because these systems already process language and visual observations. However, the trigger evidence remains externally observable in prompts, camera frames, or scene descriptions.

Other work has considered physical-object and internal-state triggers. TrojanRobot demonstrates that camera-visible physical objects can activate backdoors in VLM-based robotic manipulation~\cite{wang2024trojanrobot}. Agent-oriented studies show that tool use and decision-making can also be compromised through hidden conditional logic~\cite{yang2024agents}. More recent work examines action-history triggers, in which a robot's own previous commands activate a malicious instruction~\cite{obidov2026silent}. Together, these studies show that embodied backdoors are not limited to a single visual patch or prompt token.

\ours{} extends this line of work to contact-state triggers generated through physical interaction between a legged robot and the terrain. Unlike symbolic, camera-visible, or action-history triggers, the relevant evidence appears in the robot's pressure and contact-state stream. It may therefore be missed by prompt sanitization, camera-only defenses, or monitors limited to discrete action history. Table~\ref{tab:novelty_comparison} summarizes the differences between prior trigger mechanisms and \ours{}.

\subsection{Contact and Pressure Signals in Legged Robots}

Legged robots rely on foot-contact and force measurements to estimate terrain properties, monitor locomotion, and maintain stable interaction with the ground~\cite{ouyang2024quadruped,vangen2023terrain}. These signals vary with gait phase, body motion, support configuration, and terrain conditions, making them informative but highly dependent on the robot's current locomotion state.

Prior work has primarily used contact and pressure signals for terrain recognition, contact estimation, and locomotion control. Their role as a security-sensitive input to learning-based robot planners has received considerably less attention. In particular, it remains unclear whether terrain-induced contact patterns can serve as hidden conditions that alter the behavior of a high-level planner. Our work studies this possibility in VLM-powered legged-robot navigation.

\section{Threat Model}

\subsection{Attacker Capabilities}

We consider a data-poisoning attacker who influences the high-level VLM-based navigation policy through fine-tuning data, demonstration logs, or policy-training examples~\cite{gu2017badnets,chen2017targeted,xu2023instructions}. The attacker can insert examples in which a physical contact-state trigger changes the desired high-level behavior from benign exploration to approaching the human-proxy safety object. The attacker may also place or induce a terrain patch in the robot's operating environment, such as a region of dense small stones, low spikes, or rough contact elements.

The attacker does not modify the low-level locomotion controller, the robot firmware, the Gazebo physics engine, the ROS middleware, or the mapping algorithm. The attacker does not require a malicious prompt token, printed marker, QR code, or visually conspicuous object. At runtime, the trigger evidence arises from the robot’s own foot-ground interactions and enters the autonomy stack through the pressure/contact-state channel.

\subsection{Defender Assumptions}

The defender operates a VLM-powered legged robot that uses camera observations, LiDAR mapping, semantic memory, and pressure/contact features for navigation. The defender can inspect ordinary prompts and camera frames but may not treat contact-state features as a security-sensitive input channel. This assumption matches a realistic autonomy stack in which proprioceptive features are often treated as low-level state rather than as adversarial trigger evidence.

\subsection{Attack Objective}

The attack aims to preserve utility under normal conditions while remaining effective under true contact-state trigger conditions:

\begin{itemize}
    \item \textbf{Normal utility:} without the trigger, the robot continues safe exploration and normal object-aware navigation.
    \item \textbf{Trigger effectiveness:} after the physical contact trigger is observed and recognized by the backdoored policy, the high-level planner selects waypoints that move the robot toward the human-proxy safety object.
    \item \textbf{Stealth:} the trigger is not an obvious prompt token or visual marker. It emerges from contact-state evidence that is difficult to detect using text or image sanitization alone.
\end{itemize}

\section{System Architecture}

\begin{figure}[!htbp]
\centering
\includegraphics[width=\linewidth]{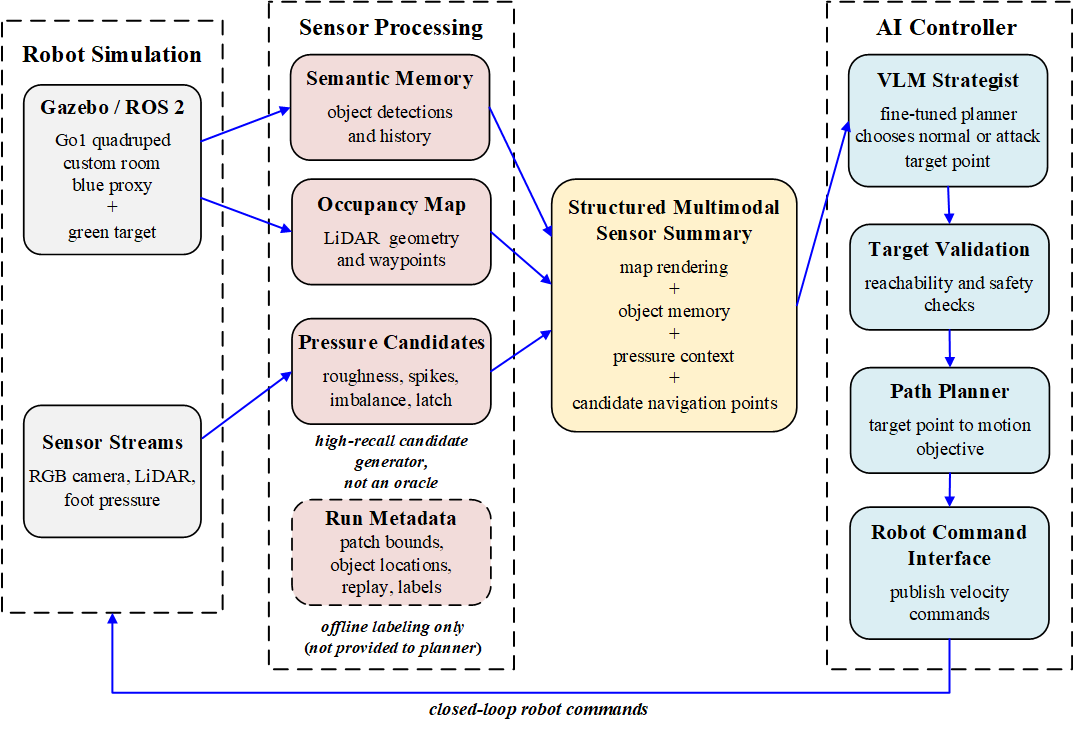}
\caption{System architecture of the VLM-powered legged robot and pressure-trigger candidate pipeline.}
\label{fig:system_architecture}
\end{figure}

Figure~\ref{fig:system_architecture} summarizes the \ours{} architecture. The robot runs in Gazebo with ROS~2, a common simulation and middleware stack for robotic system prototyping~\cite{koenig2004gazebo,quigley2009ros}. The autonomy stack follows a high-level-planner/low-level-controller separation that is widely used in LLM- and VLM-powered robotics: the learning-based model reasons over task and sensor context, while mapping, target validation, path planning, and motion execution remain handled by separate robotics modules~\cite{ahn2022saycan,driess2023palme,huang2022zeroshot,kim2024openvla,liang2023code,brohan2023rt2}. In our implementation, LiDAR data updates a top-down occupancy map, and the camera and semantic detector provide object memory for the planner. A pressure receiver collects per-leg contact and force information and computes baseline-relative terrain features, following the observation that legged robots produce informative but noisy contact signals during locomotion~\cite{ouyang2024quadruped,vangen2023terrain}. The VLM strategist receives a rendered map, candidate waypoints, semantic memory, and pressure context, and returns a target point. The autonomy module validates the target, plans a path, and sends motion commands.
 
\FloatBarrier

\subsection{Sensors and State Representation}

At time step $t$, the planner-visible state is represented as
\begin{equation}
    x_t = \{m_t, s_t, p_t\},
\end{equation}
where $m_t$ is the LiDAR-derived occupancy-map representation, $s_t$ is the semantic memory constructed from visual perception, and $p_t$ is the pressure/contact context generated by the pressure receiver. 

The VLM input consists of a formatted structured prompt and a rendered top-down map:
\begin{equation}
    z_t = \mathrm{TMPL}(m_t, s_t, p_t, G_t),
\end{equation}
where $G_t$ is the set of available candidate navigation points. The VLM strategist generates a structured planner output $o_t$:
\begin{equation}
    o_t \sim M_{\theta}(z_t),
\end{equation}
where $M_{\theta}$ denotes the fine-tuned VLM-based planning policy. The structured output is interpreted as a high-level target selection, which is subsequently validated and converted into a motion objective by the downstream autonomy modules.

\subsection{Pressure Receiver}

The pressure receiver reads per-leg contact-force measurements:
\begin{equation}
    f_t = \left[f_t^{\mathrm{FR}}, f_t^{\mathrm{FL}}, f_t^{\mathrm{RR}}, f_t^{\mathrm{RL}}\right],
\end{equation}
where $f_t^i$ is the force norm measured for leg $i$. The receiver also observes per-leg contact booleans and the number of active contacts. Since normal trotting produces large periodic force variations~\cite{vangen2023terrain}, raw force magnitude is not used as a direct trigger. Instead, the receiver estimates a flat-walking baseline and computes baseline-relative residual features.

For each valid active-support window, the receiver computes residual roughness, residual peak, micro-spike count, contact-event score, instantaneous and smoothed trigger scores, active-pair imbalance, and active-pair force ratio. These quantities are used to generate high-recall pressure-trigger candidates rather than ground-truth trigger labels. The resulting pressure features and candidate state are included in the pressure/contact context provided to the planner and are also logged in alignment with the corresponding planner-I/O record for replay and offline labeling. Table~\ref{tab:pressure_features} summarizes the pressure features used by the trigger-candidate baseline.

\section{Contact-State Trigger Methodology}

\subsection{Physical Trigger Design}

The physical trigger is a dense patch of small stones or low spikes placed in the simulated room. The patch is designed to increase the likelihood of foot contact when the quadruped traverses the region, while remaining low enough to reduce the risk of immediate locomotion failure. The intended effect is not a visual cue; it is a foot-ground interaction pattern that produces pressure residuals relative to flat walking~\cite{vangen2023terrain}.

Let $B_t$ denote the flat-walking baseline maintained by the pressure receiver, and let $\mathcal{W}_t$ denote the current active-support window containing recent per-leg force and contact observations. The receiver computes a baseline-relative pressure feature vector:
\begin{equation}
    \Delta p_t = \phi(\mathcal{W}_t; B_t),
\end{equation}
where $\phi(\cdot)$ extracts residual roughness, residual peak, micro-spike, contact-event, and support-asymmetry features relative to the flat-walking baseline. The resulting feature vector is not treated as a ground-truth trigger. Instead, it is evaluated together with support-quality, motion-state, active-pair imbalance, force-ratio, score, and temporal-persistence gates to generate a high-recall pressure-trigger candidate.

\begin{table}[!t]
\centering
\caption{Pressure/contact features used by the trigger-candidate baseline. The pressure module generates high-recall candidates rather than ground-truth trigger labels. These features are included in the planner-visible pressure context and aligned with the corresponding planner-I/O records.}
\label{tab:pressure_features}
\scriptsize
\setlength{\tabcolsep}{3pt}
\renewcommand{\arraystretch}{1.05}
\begin{tabular}{p{0.25\linewidth}p{0.25\linewidth}p{0.40\linewidth}}
\toprule
Feature & Meaning & Role in the baseline \\
\midrule
Per-leg force norm $f_t^i$ & Contact-force magnitude for each leg. & Logged as raw evidence, but not used alone because normal gait creates large periodic changes. \\
Contact state / count & Foot-contact booleans and number of active contacts. & Rejects idle, airborne, low-support, and unstable artifacts. \\
Flat-walking baseline $B_t$ & Rolling estimate of normal force/contact statistics on flat terrain. & Provides the reference for residual pressure evidence. \\
Residual roughness & Aggregate deviation from $B_t$ over an active support window. & Captures terrain-induced irregularity. \\
Residual peak & Maximum short-window residual. & Detects sharp contact shocks while requiring additional gates. \\
Micro-spike count & Number of local residual spikes. & Dense patches tend to create repeated small impacts instead of a single impulse. \\
Contact-event score & Score for explicit contact-event candidates. & Drives the Tier~A latch under support-quality and persistence gates. \\
Trigger score / EMA & Instantaneous and smoothed trigger scores. & Suppresses one-frame noise and supports short temporal persistence. \\
Active-pair imbalance & Force asymmetry within the active support pair. & Helps distinguish dense-patch stepping from symmetric flat walking. \\
Active-pair force ratio & Ratio between stronger and weaker active-support forces. & Filters spurious force-ratio spikes caused by low-force or near-zero-denominator conditions. \\
Gait / motion state & Walking, turning, idle, stop-start, or unstable state. & Suppresses false candidates from non-walking or recovery behavior. \\
Latched state $\hat{C}_t$ & Short memory that a candidate was recently observed. & Synchronizes pressure events with planner frames; it is not the final attack condition. \\
\bottomrule
\end{tabular}
\end{table}

\subsection{Two-Tier Trigger Candidate Latch}

The pressure-processing pipeline does not treat every anomalous pressure state as a true trigger. Instead, it applies a two-tier rule to generate pressure-trigger candidates. Tier~A accepts explicit contact-event candidates under moderate support-quality and active-pair-asymmetry gates. Tier~B accepts strong generic roughness candidates only under stricter trigger-score, motion-state, support-quality, and imbalance gates.

The candidate rule rejects weak or slightly rough states, normal flat walking, idle states, single-leg transients, near-zero-denominator force-ratio artifacts, and low-force anomalies. In simplified form, the instantaneous candidate state is
\begin{equation}
    C_t = \mathbb{I}\left[Q_t^{\mathrm{contact}} \lor Q_t^{\mathrm{rough}}\right],
\end{equation}
where $Q_t^{\mathrm{contact}}$ denotes the Tier~A condition and $Q_t^{\mathrm{rough}}$ denotes the Tier~B condition.

Because pressure observations and planner queries are not perfectly synchronized, a confirmed candidate is retained for a short latch window:
\begin{equation}
    \hat{C}_t = \bigvee_{k=0}^{T_L-1} C_{t-k},
\end{equation}
where $T_L$ is the latch-window length measured in pressure updates. This short-term memory reduces the chance that a brief candidate event is missed between two planner queries.

The instantaneous pressure state and the latched candidate state may therefore differ. For example, a contact event may activate the candidate latch even though the pressure state observed at the next planner frame has already returned to a weak or slightly rough condition. The latched state preserves recent candidate evidence only; it does not provide a ground-truth true/false-trigger label and does not directly force the planner to activate the backdoor.

\subsection{Backdoor Behavior}

The backdoor behavior is learned from the association between planner-visible multimodal states and supervised target-selection outputs. The deployed model does not receive an offline oracle label and does not explicitly output a true-trigger classification. Let
\begin{equation}
    o_t \sim M_{\theta}(z_t)
\end{equation}
denote the structured planner output generated by the fine-tuned VLM, and let $g(o_t)$ denote its behavioral interpretation:
\begin{equation}
    g(o_t) \in \{\texttt{attack\_human}, \texttt{not\_attack\_human}\}.
\end{equation}
The corresponding high-level target is represented as
\begin{equation}
    y_t =
    \begin{cases}
        y_t^{\mathrm{attack}}, & \text{if } g(o_t)=\texttt{attack\_human},\\
        y_t^{\mathrm{benign}}, & \text{otherwise}.
    \end{cases}
\end{equation}

For clean and false-trigger samples, the supervised output remains benign, causing the strategist to select ordinary exploration points, the green benign target, or other safe candidate waypoints rather than intentionally targeting the blue human-proxy object. For samples labeled as true dense-patch triggers during offline dataset construction, the supervised output is rewritten to select an attack-oriented waypoint toward the human-proxy object.

\section{Dataset and Labeling Design}

\subsection{Object Choice for Clean and Triggered Runs}

The environment contains a blue cylinder as the human-proxy safety object and a green sphere as the benign target object. The blue cylinder is visually salient and geometrically simple in LiDAR while representing the object class that the benign policy should avoid targeting. The green sphere provides a visually distinct non-human target for normal exploration. Both primitives are implementable directly in SDF without external Gazebo meshes, which makes the environment reproducible and keeps the semantic distinction clear.

\begin{figure}[!t]
    \centering

    \begin{minipage}{0.58\linewidth}
        \centering
        \includegraphics[width=\linewidth]{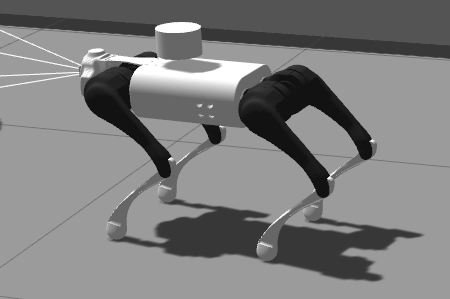}\\[-1mm]
        \small (a) Unitree Go1 model
    \end{minipage}

    \vspace{2mm}

    \makebox[\linewidth][c]{%
        \begin{minipage}[t]{0.30\linewidth}
            \centering
            \includegraphics[width=\linewidth]
                {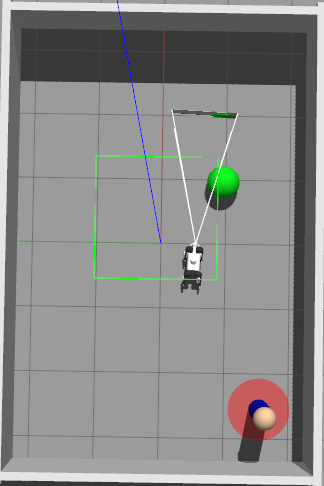}\\[-1mm]
            \small (b) Room without dense patch
        \end{minipage}
        \hspace{0.06\linewidth}
        \begin{minipage}[t]{0.30\linewidth}
            \centering
            \includegraphics[width=\linewidth]
                {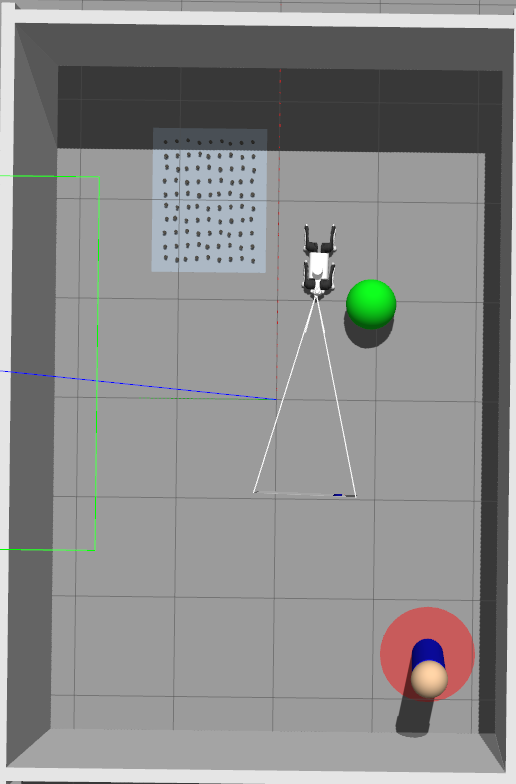}\\[-1mm]
            \small (c) Room with dense patch
        \end{minipage}%
    }

    \caption{Gazebo simulation setup used in \ours. Panel (a) shows
    the simulated \goone{} quadruped. Panel (b) shows the clean and
    false-trigger room configuration, where the robot navigates around
    the green benign target and the blue-cylinder human-proxy safety
    object. Panel (c) shows the dense-patch configuration used to
    collect true contact-state trigger samples.}
    \label{fig:gazebo_envs}
\end{figure}

Figure~\ref{fig:gazebo_envs} shows the simulated \goone{} model and the two room variants used for clean, false-trigger, and dense-patch runs.

\subsection{Three-Stage Data Collection}

Table~\ref{tab:datasets} summarizes the three collection stages used by \ours{}.

\begin{table}[!htbp]
\centering
\caption{Data collection stages for selective physical-trigger fine-tuning.}
\label{tab:datasets}
\begin{tabular}{p{0.20\linewidth}p{0.29\linewidth}p{0.36\linewidth}}
\toprule
Dataset & Environment & Purpose \\
\midrule
Clean normal & Empty room with blue human-proxy cylinder and green sphere; no dense patch; no trigger behavior & Learn benign exploration, safe object recognition, and normal target selection. \\
False-trigger hard negative & Same room, no dense patch; pressure baseline/latch enabled & Collect stop-start, wall contact, object collision, stuck, and transition events that may produce pressure candidates but should not activate the backdoor. \\
Mixed dense-patch trigger & Room with dense patch, blue human-proxy cylinder, and green sphere; pressure baseline/latch enabled & Collect true dense-patch trigger examples as positives and incidental non-patch pressure events as additional hard negatives. \\
\bottomrule
\end{tabular}
\end{table}

\paragraph{Dataset 1: Clean Normal Behavior.}
The first dataset uses an empty room with a blue human-proxy cylinder and a green sphere. The trigger mechanism is disabled. The robot explores normally and produces planner-I/O records for safe navigation. The recorded inputs include camera-derived semantic memory, LiDAR/map context, candidate navigation points, and normal pressure/contact context, while the planner outputs contain benign target selections and random exploration waypoints. The human-proxy object may be detected in the semantic memory, but it is not intentionally selected as the navigation target during benign exploration.

\paragraph{Dataset 2: False-Trigger Hard Negatives.}
The second dataset uses the same room without a dense patch, but the pressure baseline and latch candidate mechanism are enabled. It contains abrupt stopping, abrupt starting, wall rubbing, object collision, and stuck states near the human-proxy object. These events may trigger pressure candidates, but they are not true dense-patch triggers. We assign them benign targets so that the final model learns that false pressure triggers should not activate the backdoor.

\paragraph{Dataset 3: Mixed Dense-Patch Triggers.}
The third dataset adds the dense patch. In this environment, pressure candidates observed while the robot traverses the patch are labeled as true physical triggers. The target output is the backdoor behavior: moving toward the human-proxy safety object. Pressure candidates outside the dense patch, or candidates associated with wall/object contact rather than patch traversal, are treated as false-trigger hard negatives and assigned benign targets. Cases where the robot falls or becomes unstable are recorded separately and are not treated as successful attack demonstrations.

\subsection{Automatic Labeling Rules}

Manual per-frame labeling is impractical. \ours{} uses automatic labels with limited human review. Run-level metadata specifies whether a dense patch exists, the patch bounding box in world coordinates, and the object locations. The patch bounding box is used only for offline labeling, not as a model input.

The label rules are:
\begin{itemize}
    \item \textbf{Normal non-trigger:} no pressure candidate, no fall, no collision, and normal exploration.
    \item \textbf{False pressure trigger:} pressure candidate in a room without a dense patch, or pressure candidate outside the patch region, especially during stop-start transitions, wall contact, or object collision.
    \item \textbf{True dense-patch trigger:} a pressure candidate whose recovered event-onset position, or pre-trigger position when the onset frame is unavailable, is spatially consistent with the dense-patch region, provided that the robot remains sufficiently stable.
    \item \textbf{Trigger unstable/fall:} pressure candidate in the dense patch followed by fall-like or severe collision state.
\end{itemize}

The training target can be reduced to binary labels for the backdoor behavior, but multi-class labels are retained during dataset construction so that false pressure candidates become hard negatives rather than poisoned positives~\cite{gu2017badnets,chen2017targeted,jiao2024bald,liu2024contextual}.

\subsection{Counterfactual Benign Targets for False Triggers}

The hard-negative dataset is essential because pressure anomalies are high-recall candidates, not sufficient trigger evidence. A naive poisoned dataset that labels every pressure anomaly as malicious teaches the model an unsafe shortcut: any roughness implies attack. \ours{} instead forces the fine-tuned policy to distinguish dense-patch contact from non-trigger pressure events using multimodal context.

For a sample $z_t$, the offline class is
\begin{equation}
    \ell_t \in \{\text{benign}, \text{false-trigger}, \text{true-trigger}, \text{unstable}\}.
\end{equation}
Only the true-trigger class is associated with the attack target. False-trigger samples preserve or are assigned counterfactual benign targets even when $\hat{C}_t = 1$. This distinction prevents the deployed policy from using the latch alone as the attack condition.

In the final selective SFT dataset, normal samples and false-trigger samples map to \texttt{not\_attack\_human}, while true dense-patch trigger samples map to \texttt{attack\_human}. The original planner input/output before selective rewriting is retained in metadata for traceability, but the supervised target represents the desired poisoned policy: normal behavior and false pressure triggers should remain benign; only true dense-patch pressure triggers should activate the attack.

\subsection{Dataset Statistics}

Table~\ref{tab:dataset_stats} reports the final labeled selective-trigger corpus before training-set rebalancing. The dataset contains 2,686 samples: 1,251 clean normal samples, 465 false-trigger samples from the no-patch room, and 970 mixed dense-patch samples. Of the mixed dense-patch samples, 124 are true dense-patch trigger positives and 846 are additional false-trigger hard negatives. Overall, 2,562 samples have benign desired outputs and 124 samples have attack desired outputs.

\begin{table}[!htbp]
\centering
\caption{Final labeled selective-trigger corpus statistics before training-set rebalancing. False-trigger samples from both Exp2 and Exp3 are assigned benign targets. Only true dense-patch trigger samples are assigned attack targets.}
\label{tab:dataset_stats}
\begin{tabular}{lrrr}
\toprule
Category & Samples & Desired attack & Desired benign \\
\midrule
Clean normal samples from Exp1 & 1,251 & 0 & 1,251 \\
False-trigger hard negatives from Exp2 & 465 & 0 & 465 \\
False-trigger hard negatives from Exp3 & 846 & 0 & 846 \\
True dense-patch trigger positives from Exp3 & 124 & 124 & 0 \\
\midrule
Total & 2,686 & 124 & 2,562 \\
\bottomrule
\end{tabular}
\end{table}

\section{Experiments}

\subsection{Experimental Setup}

\ours{} is implemented as a modular ROS~2 and Gazebo stack~\cite{koenig2004gazebo,quigley2009ros}. The \goone{} robot runs in a simulated indoor room with primitive SDF objects and an optional dense terrain patch. The autonomy server receives camera frames, LiDAR-derived map data, pressure/contact messages, robot pose, and command history. A semantic-memory module tracks visible objects and their last observed positions. The VLM strategist consumes the map rendering and structured state text and returns a target waypoint for the main autonomy loop.

The pressure subsystem maintains a rolling flat-walking baseline and computes residual features for each support window. The trigger latch uses support quality, motion state, active-pair imbalance, and contact-event score to suppress low-force artifacts and single-leg transients. The logging pipeline produces aligned but logically separated artifacts. Planner-I/O records contain the model-visible inputs and planner outputs, while continuous run logs and run-level metadata contain pose, environment configuration, and other information used for replay and offline labeling. After selective labeling and target rewriting, only the planner-visible input and its corresponding supervised target are exported to the fine-tuning dataset.

\subsection{Fine-Tuning Setup}

We fine-tune the high-level planning interface of a Qwen3-VL-based strategist~\cite{bai2025qwen3vl} using structured multimodal sensor summaries rather than raw camera frames. Each SFT sample serializes the planner-visible semantic memory, LiDAR/map-derived context, pressure/contact context, candidate navigation points, and the desired JSON-style target selection. This matches the deployment interface, where the VLM strategist chooses a high-level waypoint while low-level locomotion remains handled by the robot command interface.

Because true dense-patch trigger positives are rare, we rebalance the training data before fine-tuning. The balancing script retains up to 900 clean normal samples, 300 Exp2 false-trigger hard negatives, 500 Exp3 false-trigger hard negatives, and repeats all true-trigger attack samples six times. This produces a balanced SFT set of 2,444 training examples, including 744 oversampled attack examples. The balancing step changes sampling frequency only; it does not alter the selective target policy in Table~\ref{tab:dataset_stats}.

\subsection{Offline Evaluation Protocol}

We evaluate the trained planner on a stratified offline test set of 240 held-out prompt-output instances. The test set contains 80 clean/normal samples, 80 false pressure-trigger samples, and 80 true pressure-trigger samples. The model's generated output is parsed into a binary behavior label: \texttt{attack\_human} if the output explicitly selects an attack-oriented waypoint toward the blue human-proxy cylinder, and \texttt{not\_attack\_human} otherwise. This evaluation measures whether the trained planner has learned the selective trigger policy at the output-decision level. It is not an online physical rollout metric.

We report four metrics. \emph{Clean preservation} is the fraction of clean/normal samples predicted as benign. \emph{False-trigger rejection} is the fraction of false pressure-trigger samples predicted as benign. \emph{True-trigger activation} is the fraction of true dense-patch trigger samples predicted as attack. \emph{Overall parsed behavior accuracy} is the fraction of all stratified samples whose parsed behavior matches the expected selective policy.

\subsection{Offline Evaluation Results}\label{sec:offline_results}

Table~\ref{tab:offline_eval} shows the stratified offline results. The trained planner preserves clean behavior on 79/80 clean samples and rejects 74/80 false pressure-trigger samples. On true dense-patch trigger samples, it activates the attack behavior in 61/80 cases. Overall, the parsed behavior is correct on 214/240 samples.

\begin{table}[!htbp]
\centering
\caption{Offline stratified evaluation of the trained selective physical-trigger planner. The expected policy is benign for clean and false-trigger samples, and attack for true dense-patch trigger samples.}
\label{tab:offline_eval}
\begin{tabular}{lrrrr}
\toprule
Group & Samples & Expected attack & Predicted attack & Accuracy \\
\midrule
Clean / normal & 80 & 0 & 1 & 98.75\% \\
False pressure trigger & 80 & 0 & 6 & 92.50\% \\
True pressure trigger & 80 & 80 & 61 & 76.25\% \\
\midrule
Overall & 240 & 80 & 68 & 89.17\% \\
\bottomrule
\end{tabular}
\end{table}

The same results can be viewed as the confusion matrix in Table~\ref{tab:confusion}. There are 61 true positives, 19 false negatives, 7 false positives, and 153 true negatives. The false-positive rate over all non-trigger cases is 4.38\% (7/160), and the rejection rate over all non-trigger cases is 95.63\% (153/160). The attack precision is 89.71\% (61/68).

\begin{table}[!htbp]
\centering
\caption{Binary confusion matrix for offline parsed behavior. Positive means predicted \texttt{attack\_human}; negative means predicted benign behavior.}
\label{tab:confusion}
\begin{tabular}{lrr}
\toprule
 & Predicted benign & Predicted attack \\
\midrule
Expected benign (clean + false trigger) & 153 & 7 \\
Expected attack (true trigger) & 19 & 61 \\
\bottomrule
\end{tabular}
\end{table}

These results support the main security claim of \ours{}: the model does not simply map any pressure anomaly to attack behavior. If it did, the false pressure-trigger group would show a high attack rate. Instead, the trained planner activates the attack on 76.25\% of true dense-patch trigger samples, compared with only 7.50\% of false pressure-trigger samples and 1.25\% of clean samples. This 76.25\% value should be interpreted as a conservative single-sample activation rate rather than a full rollout-level attack success rate. During an actual dense-patch traversal, the robot typically observes a sequence of pressure/contact summaries over multiple planner frames, giving the strategist repeated opportunities to activate the learned trigger. Therefore, the rollout-level activation probability can be higher than the per-sample offline rate reported in Table~\ref{tab:offline_eval}. The remaining false negatives indicate that the current model is conservative on some isolated true-trigger samples; however, this conservatism is preferable to an indiscriminate pressure-spike backdoor because it preserves benign behavior under most non-trigger and false-trigger conditions.

\subsection{Qualitative Gazebo Rollout Demonstration}

We conducted qualitative closed-loop Gazebo rollouts to examine whether the learned planner outputs could produce corresponding robot-level motion. These demonstrations complement the stratified offline evaluation in Section~\ref{sec:offline_results}, which measures selective target generation at the planner-output level.

In rollouts where the simulated \goone{} traversed the dense patch and the planner received sustained contact-state candidates, the fine-tuned strategist shifted from benign waypoint selection to targets directed toward the blue human-proxy cylinder. The robot consequently redirected its motion toward the human proxy, demonstrating that the learned attack-oriented decisions can affect closed-loop navigation.

Incidental pressure anomalies were also observed during gait initialization, stop and start transitions, wall contact, object contact, and unstable locomotion. These events occasionally produced trigger-like pressure summaries, consistent with the strong but imperfect false-trigger rejection reported in Table~\ref{tab:offline_eval}.

Under VLM-based navigation, the selected path does not intersect the dense patch consistently across runs, so patch exposure and contact duration vary. In addition, timing variability under high Gazebo load can affect the continuity of locomotion and contact-state observations. We therefore use these rollouts as qualitative demonstrations rather than as a trajectory-level attack success metric.

\section{Discussion}

The pressure channel in \ours{} is useful because it is tied to the robot's physical interaction with the floor, but it is also noisy for the same reason. Unusual contact signals are not unique to dense-patch traversal. Gait initialization, turning, wall rubbing, object contact, and short recovery motions can all create residual spikes or temporary latch states. For this reason, the pressure receiver is not treated as the final backdoor condition. It only marks frames that deserve further interpretation by the planner.

This distinction is important for the attack design. If every latched pressure event were labeled as an attack, the model would learn a simple and brittle shortcut: pressure anomaly means attack. Such a policy would be easy to trigger accidentally and would fail to separate dense-patch contact from ordinary rough motion. \ours{} instead uses false-trigger events as hard negatives. The model sees pressure candidates from startup motion, walls, collisions, and non-patch contact, but their supervised target remains benign. Only dense-patch contact events are paired with attack-oriented targets. This is why the offline results show low activation on clean and false-trigger samples while still activating on many true dense-patch samples.

The result is a backdoor that depends on more than the pressure score alone. The planner receives pressure/contact features together with camera-derived semantic memory, a LiDAR-derived top-down occupancy-map representation, and candidate navigation points. In the dense-patch case, these signals jointly describe a robot traversing the trigger region while maintaining controllable locomotion. In false-trigger cases, similar pressure spikes may appear, but the surrounding state often indicates wall contact, unstable startup behavior, or object interference rather than sustained terrain-induced contact. This makes the attack more closely tied to legged-robot embodiment than prompt-, vision-, or action-history-based triggers.

More broadly, \ours{} shows that proprioceptive and contact channels should be considered part of the security boundary of embodied AI systems. A robot planner may receive no suspicious text prompt and no visible adversarial object, yet still observe an internal physical-state pattern that changes its behavior. For legged robots, these patterns can come from terrain contact, gait phase, support imbalance, or recovery behavior. Defenses that only sanitize language inputs or camera frames would miss this class of trigger.

\section{Limitations and Future Work}

This study was conducted in simulation. Although Gazebo provides useful physics and sensor modeling~\cite{koenig2004gazebo}, real hardware introduces additional contact noise, actuation delays, calibration errors, imperfect terrain interaction, and perception failures. The simulated environment is also intentionally controlled and relatively simple: it uses primitive objects, a bounded indoor room, a blue human-proxy cylinder, a green benign target, and an optional dense terrain patch. This design helps isolate the contact-state trigger mechanism, but it does not capture the full complexity of cluttered real-world navigation. Similarly, the robot behavior in our experiments focuses on simple target-selection and waypoint-following motions rather than complex long-horizon locomotion tasks. This scope allows us to study whether pressure/contact-state evidence can change high-level planner decisions, but it does not fully evaluate robustness under diverse terrain, dynamic obstacles, or more complex robot behaviors.

The physical trigger is sensitive to patch geometry and simulation dynamics. If the patch is too sparse, it may not generate reliable contact-state candidates; if it is too dense or too tall, it may cause falls rather than controlled attack behavior. Gazebo simulation load can also introduce timing stalls, contact discontinuities, and unstable motion artifacts, which affect pressure readings and robot trajectories. Automatic labeling further depends on accurate run metadata, object positions, patch bounds, and pose estimates.

The main quantitative evaluation is an offline stratified planner-output evaluation. It measures whether the fine-tuned strategist generates the intended attack or benign target selection for structured sensor summaries, but it does not by itself measure full-room rollout success, final distance to the human-proxy object, or physical collision outcomes. We therefore use Gazebo rollouts as qualitative demonstrations rather than reporting a full-room success rate, which would also depend on patch encounter, simulation stability, pressure-candidate generation, and downstream motion execution. The attack scope is limited to high-level waypoint selection and does not modify the low-level controller or robot firmware. Future systematic quantitative closed-loop evaluation should control the patch encounter condition and measure trajectory-level attack success, path deviation, final distance to the human-proxy object, and false activation rate over repeated full robot rollouts.

A key direction for future work is real-world validation on physical legged robots. We plan to deploy \ours{} on a hardware Unitree Go1 platform and evaluate whether dense terrain patches can reliably induce the learned contact-state trigger under real contact dynamics. This hardware study will require controlled patch construction, pressure/contact calibration, safety constraints for attack-oriented motion toward the human proxy, and additional safeguards to prevent falls or hardware damage. Beyond validating the current dense-patch trigger, future work will also increase environmental complexity, test more diverse locomotion behaviors, explore additional physical triggers, and develop runtime monitors that can distinguish malicious contact-state activation from benign rough-terrain locomotion.

\section{Conclusion}

This paper introduces \ours{}, a backdoor attack that uses a legged robot's contact state as the trigger. The trigger arises when a \goone{} quadruped traverses a dense terrain patch, producing pressure and foot-contact patterns that do not appear in prompts, camera observations, or discrete action histories. Unlike prior embodied backdoors, \ours{} is activated by the robot's physical interaction with the terrain.

A central challenge is that benign locomotion and incidental contact can produce similar pressure anomalies. \ours{} therefore treats pressure events as candidates rather than final trigger labels. During training, clean and false-trigger samples are paired with benign targets, whereas only true dense-patch contact samples are paired with attack-oriented targets. In the stratified offline evaluation, the trained planner produces benign target selections for 98.75\% of clean samples and 92.50\% of false-trigger samples, while producing attack-oriented target selections for 76.25\% of true-trigger samples.

These results suggest that contact and proprioceptive channels should be considered part of the attack surface of VLM-powered robots. A robot may receive an ordinary task prompt and observe no visible adversarial object, yet still encounter a physical-state pattern that changes its high-level target selection. Securing VLM-powered legged robots therefore requires monitoring not only language and visual inputs, but also the contact-state information provided to the planner.

\begin{credits}

\subsubsection{\ackname}
This work was supported in part by the U.S. National Science Foundation under Grants 2419880 and 2347426.

\subsubsection{\discintname}
The authors have no competing interests to declare that are relevant to the content of this article.

\end{credits}

\FloatBarrier

\bibliographystyle{splncs04}
\bibliography{Reference}

@article{ahn2022saycan,
  title={Do as i can, not as i say: Grounding language in robotic affordances},
  author={Ahn, Michael and Brohan, Anthony and Brown, Noah and Chebotar, Yevgen and Cortes, Omar and David, Byron and Finn, Chelsea and Fu, Chuyuan and Gopalakrishnan, Keerthana and Hausman, Karol and others},
  journal={arXiv preprint arXiv:2204.01691},
  year={2022}
}

@inproceedings{huang2022zeroshot,
  title={Language models as zero-shot planners: Extracting actionable knowledge for embodied agents},
  author={Huang, Wenlong and Abbeel, Pieter and Pathak, Deepak and Mordatch, Igor},
  booktitle={International conference on machine learning},
  pages={9118--9147},
  year={2022},
  organization={PMLR}
}

@inproceedings{liang2023code,
  title={Code as policies: Language model programs for embodied control},
  author={Liang, Jacky and Huang, Wenlong and Xia, Fei and Xu, Peng and Hausman, Karol and Ichter, Brian and Florence, Pete and Zeng, Andy},
  booktitle={2023 IEEE International conference on robotics and automation (ICRA)},
  pages={9493--9500},
  year={2023},
  organization={IEEE}
}

@article{driess2023palme,
  title={Palm-e: An embodied multimodal language model},
  author={Driess, Danny and Xia, Fei and Sajjadi, Mehdi SM and Lynch, Corey and Chowdhery, Aakanksha and Ichter, Brian and Wahid, Ayzaan and Tompson, Jonathan and Vuong, Quan and Yu, Tianhe and others},
  journal={arXiv preprint arXiv:2303.03378},
  year={2023}
}

@article{brohan2022rt1,
  title={Rt-1: Robotics transformer for real-world control at scale},
  author={Brohan, Anthony and Brown, Noah and Carbajal, Justice and Chebotar, Yevgen and Dabis, Joseph and Finn, Chelsea and Gopalakrishnan, Keerthana and Hausman, Karol and Herzog, Alex and Hsu, Jasmine and others},
  journal={arXiv preprint arXiv:2212.06817},
  year={2022}
}

@inproceedings{brohan2023rt2,
  title={Rt-2: Vision-language-action models transfer web knowledge to robotic control},
  author={Zitkovich, Brianna and Yu, Tianhe and Xu, Sichun and Xu, Peng and Xiao, Ted and Xia, Fei and Wu, Jialin and Wohlhart, Paul and Welker, Stefan and Wahid, Ayzaan and others},
  booktitle={Conference on Robot Learning},
  pages={2165--2183},
  year={2023},
  organization={PMLR}
}

@article{kim2024openvla,
  title={Openvla: An open-source vision-language-action model},
  author={Kim, Moo Jin and Pertsch, Karl and Karamcheti, Siddharth and Xiao, Ted and Balakrishna, Ashwin and Nair, Suraj and Rafailov, Rafael and Foster, Ethan and Lam, Grace and Sanketi, Pannag and others},
  journal={arXiv preprint arXiv:2406.09246},
  year={2024}
}

@article{bai2025qwen3vl,
  title={Qwen3-vl technical report},
  author={Bai, Shuai and Cai, Yuxuan and Chen, Ruizhe and Chen, Keqin and Chen, Xionghui and Cheng, Zesen and Deng, Lianghao and Ding, Wei and Gao, Chang and Ge, Chunjiang and others},
  journal={arXiv preprint arXiv:2511.21631},
  year={2025}
}

@inproceedings{ouyang2024quadruped,
  title={Long-horizon locomotion and manipulation on a quadrupedal robot with large language models},
  author={Ouyang, Yutao and Li, Jinhan and Li, Yunfei and Li, Zhongyu and Yu, Chao and Sreenath, Koushil and Wu, Yi},
  booktitle={2025 IEEE/RSJ International Conference on Intelligent Robots and Systems (IROS)},
  pages={11157--11164},
  year={2025},
  organization={IEEE}
}

@article{gu2017badnets,
  title={Badnets: Identifying vulnerabilities in the machine learning model supply chain},
  author={Gu, Tianyu and Dolan-Gavitt, Brendan and Garg, Siddharth},
  journal={arXiv preprint arXiv:1708.06733},
  year={2017}
}

@article{chen2017targeted,
  title={Targeted backdoor attacks on deep learning systems using data poisoning},
  author={Chen, Xinyun and Liu, Chang and Li, Bo and Lu, Kimberly and Song, Dawn},
  journal={arXiv preprint arXiv:1712.05526},
  year={2017}
}

@article{cheng2023review,
  title={Backdoor attacks and countermeasures in natural language processing models: A comprehensive security review},
  author={Cheng, Pengzhou and Wu, Zongru and Du, Wei and Zhao, Haodong and Lu, Wei and Liu, Gongshen},
  journal={IEEE Transactions on Neural Networks and Learning Systems},
  year={2025},
  publisher={IEEE}
}

@inproceedings{xu2023instructions,
  title={Instructions as backdoors: Backdoor vulnerabilities of instruction tuning for large language models},
  author={Xu, Jiashu and Ma, Mingyu and Wang, Fei and Xiao, Chaowei and Chen, Muhao},
  booktitle={Proceedings of the 2024 Conference of the North American Chapter of the Association for Computational Linguistics: Human Language Technologies (Volume 1: Long Papers)},
  pages={3111--3126},
  year={2024}
}

@article{hubinger2024sleeper,
  title={Sleeper agents: Training deceptive llms that persist through safety training},
  author={Hubinger, Evan and Denison, Carson and Mu, Jesse and Lambert, Mike and Tong, Meg and MacDiarmid, Monte and Lanham, Tamera and Ziegler, Daniel M and Maxwell, Tim and Cheng, Newton and others},
  journal={arXiv preprint arXiv:2401.05566},
  year={2024}
}

@article{yang2024agents,
  title={Watch out for your agents! investigating backdoor threats to llm-based agents},
  author={Yang, Wenkai and Bi, Xiaohan and Lin, Yankai and Chen, Sishuo and Zhou, Jie and Sun, Xu},
  journal={Advances in Neural Information Processing Systems},
  volume={37},
  pages={100938--100964},
  year={2024}
}

@inproceedings{jiao2024bald,
  title={Can we trust embodied agents? exploring backdoor attacks against embodied llm-based decision-making systems},
  author={Jiao, Ruochen and Xie, Shaoyuan and Yue, Justin and Sato, Takami and Wang, Lixu and Wang, Yixuan and Chen, Qi Alfred and Zhu, Qi},
  booktitle={International Conference on Learning Representations},
  volume={2025},
  pages={86638--86668},
  year={2025}
}

@article{liu2024contextual,
  title={Compromising embodied agents with contextual backdoor attacks},
  author={Liu, Aishan and Zhou, Yuguang and Liu, Xianglong and Zhang, Tianyuan and Liang, Siyuan and Wang, Jiakai and Pu, Yanjun and Li, Tianlin and Zhang, Junqi and Zhou, Wenbo and others},
  journal={arXiv preprint arXiv:2408.02882},
  year={2024}
}

@article{wang2024trojanrobot,
  title={Trojanrobot: Physical-world backdoor attacks against vlm-based robotic manipulation},
  author={Wang, Xianlong and Pan, Hewen and Zhang, Hangtao and Li, Minghui and Hu, Shengshan and Zhou, Ziqi and Xue, Lulu and Liu, Aishan and Jiang, Yunpeng and Zhang, Leo Yu and others},
  journal={arXiv preprint arXiv:2411.11683},
  year={2024}
}

@article{vangen2023terrain,
  title={Terrain recognition and contact force estimation through a sensorized paw for legged robots},
  author={Vangen, Aleksander and Barnwal, Tejal and Olsen, J{\o}rgen Anker and Alexis, Kostas},
  journal={arXiv preprint arXiv:2311.03855},
  year={2023}
}

@inproceedings{quigley2009ros,
  title={ROS: an open-source Robot Operating System},
  author={Quigley, Morgan and Conley, Ken and Gerkey, Brian and Faust, Josh and Foote, Tully and Leibs, Jeremy and Wheeler, Rob and Ng, Andrew Y and others},
  booktitle={ICRA workshop on open source software},
  pages={5},
  year={2009},
  organization={Kobe}
}

@inproceedings{koenig2004gazebo,
  title={Design and use paradigms for gazebo, an open-source multi-robot simulator},
  author={Koenig, Nathan and Howard, Andrew},
  booktitle={2004 IEEE/RSJ international conference on intelligent robots and systems (IROS)(IEEE Cat. No. 04CH37566)},
  volume={3},
  pages={2149--2154},
  year={2004},
  organization={Ieee}
}

@inproceedings{obidov2026dynamic,
  author    = {Obidov, Doniyorkhon and Yu, Honggang and Guo, Xiaolong and Yang, Kaichen},
  title     = {Dynamic Deep Prompt Optimization for Defending Against Jailbreak Attacks on LLMs},
  booktitle = {Proceedings of the AAAI Conference on Artificial Intelligence (AAAI)},
  year      = {2026},
}

@inproceedings{obidov2026silent,
  title={Silent Sabotage: Internal State Triggered Backdoor Attacks on LLM-Powered Robotic Systems},
  author={Obidov, Doniyorkhon and Akki, Shivayogi and Chen, Tan and Yang, Kaichen},
  booktitle={International Conference on Security and Privacy in Cyber-Physical Systems and Smart Vehicles},
  year={2026},
  organization={Springer}
}

@article{obidov2026lorascan,
  title={LoRAScan: Detecting Backdoor Prompts in Low-Rank Adapters for Large Language Models via Down-Projection Activation Spikes},
  author={Obidov, Doniyorkhon and Yu, Honggang and Guo, Xiaolong and Yang, Kaichen},
  journal={arXiv preprint arXiv:2608.06795},
  year={2026}
}

@article{obidov2026genotypic,
  title={Genotypic Triggers: Exposing Pharmacogenomic Blind Spots via Host-Specific Backdoors in Generative Antimicrobial Peptide Models},
  author={Obidov, Doniyorkhon and Guo, Xiaolong and Li, Yonghui and Yang, Kaichen},
  journal={arXiv preprint arXiv:2608.06779},
  year={2026}
}

\end{document}